\documentclass{article}

     \usepackage[preprint]{neurips_2019}

\usepackage[utf8]{inputenc} 
\usepackage[T1]{fontenc}    
\usepackage[draft]{hyperref}       
\usepackage{url}            
\usepackage{booktabs}       
\usepackage{amsfonts}       
\usepackage{nicefrac}       
\usepackage{microtype}      

\usepackage{amsmath,amsfonts,bm}

\usepackage{graphicx} 
\newcommand{\real}{\mathbb{R}}
\usepackage{xcolor}

\title{A Two-Step Graph Convolutional Decoder\\ for Molecule Generation}

\author{%
  Xavier Bresson \\
  School of Computer Science and Engineering\\
  NTU, Singapore\\
  \texttt{xbresson@ntu.edu.sg} \\
  \And
  Thomas Laurent \\
  Department of Mathematics\\
  Loyola Marymount University\\
  \texttt{tlaurent@lmu.edu} \\
}

\begin{document}

\maketitle

\begin{abstract}
We propose a simple auto-encoder framework for molecule generation. The molecular graph is first encoded into a continuous latent representation $z$, which is then decoded back to a molecule. The encoding process is easy, but the decoding process remains challenging. In this work, we introduce a simple two-step decoding process. In a first step, a fully connected neural network uses the latent vector $z$ to produce a molecular formula, for example CO$_2$ (one carbon and two oxygen atoms). In a second step, a graph convolutional neural network uses the same latent vector $z$ to  place bonds between the atoms that were produced in the first step (for example a double bond will be placed between the carbon and each of the oxygens). This two-step process, in which a bag of atoms is first generated, and then assembled, provides a simple framework that allows us to develop an efficient molecule auto-encoder. Numerical experiments on basic tasks such as novelty, uniqueness, validity and optimized chemical property for the 250k ZINC molecules demonstrate the performances of the proposed system. Particularly, we achieve the highest reconstruction rate of 90.5\%, improving the previous rate of 76.7\%. We also report the best property improvement results when optimization is constrained by the molecular distance between the original and generated molecules. 
\end{abstract}

\section{Introduction}

A fundamental problem in drug discovery and material science is to design molecules with arbitrary optimized chemical properties. This is a highly challenging mathematical and computational problem. Molecules are intrinsically combinatorial. Any small perturbation in the chemical structure may result in large variation in the desired molecular property. Besides, the space of valid molecules is huge as the number of combinatorial permutations of atoms and bonds grows factorially. A common example is the space of drug-like molecules that is estimated between $10^{23}$ and $10^{60}$ in \cite{art:Polishchuk13}. Currently, most drugs are hand-crafting by years of trial-and-error by human experts in chemistry and pharmacology. The recent advances of machine learning and deep learning has opened a new research direction, with the promise to learn these molecular spaces for optimized molecule generation without hand-crafting them.

{\bf Molecule auto-encoder.} In this work we propose a simple auto-encoder for molecule generation. Each molecule is represented by a graph whose vertices correspond to the atoms and whose edges correspond to the bonds. The encoder takes as input a molecule with $N$ atoms ($N$ changes from molecule to molecule) and generates a continuous latent vector $z$ of fixed size $d$. The decoder takes as input the latent vector $z$ and attempts to recreate the molecule.

{\bf Molecule encoder with graph neural networks.} GNNs \cite{art:Scarselli08,art:SukhbaatarSzlamFergus16ComAgents,kipf2016semi,hamilton2017inductive,monti2017geometric} provide a natural way to encode molecules of varying size $N$ into vectors of fixed size $d$. GNNs have been used in \cite{duvenaud2015convolutional,gilmer2017neural} to encode molecules into $z$ and use the latent vector for regression  such as the prediction of molecule energies. Designing a good encoder is quite straightforward and can be done in two steps:
\begin{eqnarray}
x_i &= f_\textrm{node}(\{x_j\}_{j\in\mathcal{N}(i)}),\\
z &= g_\textrm{graph}(\{x_i\}_{i\in V}),
\end{eqnarray}
where $\mathcal{N}(i)$ is the neighborhood of node $i$ and $V$ is the set of nodes. These formula determines respectively a feature representation for all nodes and a latent representation for the molecular graph. Function $f_\textrm{node}$ instantiates the type of GNNs such as graph recurrent neural networks \cite{art:Scarselli08,li2015gated}, graph convolutional neural networks \cite{art:SukhbaatarSzlamFergus16ComAgents,kipf2016semi,hamilton2017inductive,monti2017geometric} or graph attention networks \cite{velivckovic2017graph}. Function $g_\textrm{graph}$ is an aggregation function of all node features, such as the $\textrm{mean}$, $\textrm{sum}$ or $\textrm{max}$. As a consequence, designing a good encoder is quite straightforward. Designing the decoder on the other hand is much more challenging.

{\bf Molecule decoder.} Two approaches have been proposed to generate the molecular graph from a latent vector. Auto-regressive models rely on generating the atoms and the bonds in a sequential way, one after the other. These models have been so far the most successful approach to generate molecules \cite{you2018graph, li2018learning, jin2018junction, kusner2017grammar}. Alternatively, one-shot models generate all atoms and and all molecules in a single pass. They have been introduced and developed in \cite{simonovsky2018graphvae,de2018molgan}. A challenge with one-shot decoder is to generate molecules of different sizes as it is hard to generate simultaneously the number of atoms and the bond structure between the atoms. \cite{simonovsky2018graphvae,de2018molgan} proposed to generate molecules with a fixed size, the size of the largest molecule in the training set. Molecules with smaller size will use special atom and bond tokens indicating none-atom and none-bond. We propose an alternative by disentangling the problems of finding simultaneously the number of atoms and the bond structure. We introduce a mechanism in which the decoder will first generate all the atoms in one shot, and then will generate all the bonds in one shot. Our one-shot decoder may not produce a chemically valid molecule as the atom valency (maximum number of electrons in the outer shell of the atom that can participate of a chemical bond) may be violated. We use greedy beam search technique to produce a valid molecule. Finally, we will formulate our system as a variational auto-encoder (VAE) for a better latent space representation of molecules.

{\bf Numerical experiments.} We evaluate the proposed molecular auto-encoder model following the experiments introduced in \cite{jin2018junction}. First, we will evaluate the reconstruction quality, novelty, uniqueness and novelty of the proposed auto-encoder. Second, we will optimize a targeted molecular property, the constrained solubility of molecules. Third, we will optimize the same chemical target while constraining the distance between the original and the generated molecules.

\section{Proposed Method}
\begin{figure*}[h!]
\centering
\includegraphics[width=14cm]{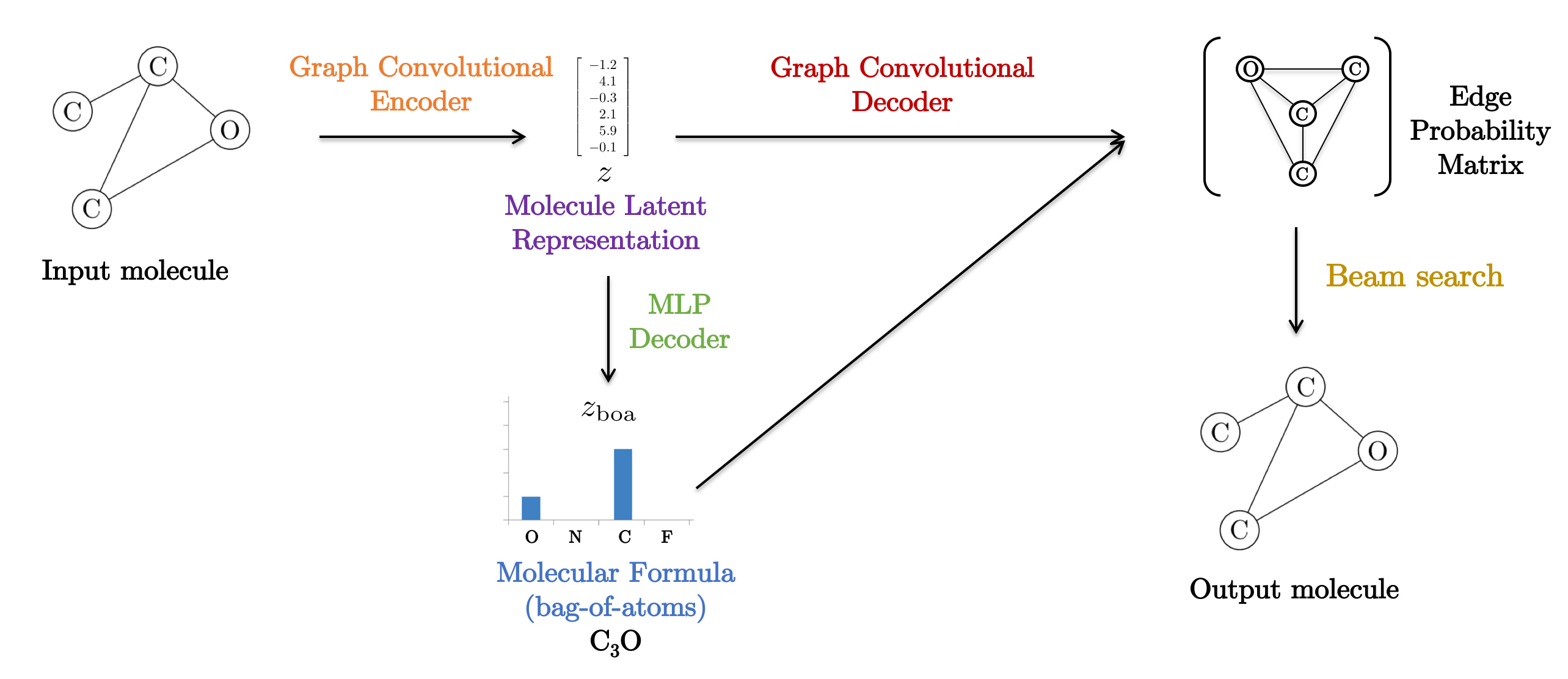}
\caption{Proposed auto-encoder. The encoder reduces the molecular graph to a latent vector $z$. The decoder uses a MLP to produce a molecular formula and a graph convolutional network classifies each bond between the atoms given by the molecular formula. Finally, a beam search generates a valid molecule.
}
\label{description}
\end{figure*}

Figure \ref{description} depicts the proposed auto-encoder. We detail each part of the system in this section.

\subsection{Molecule Encoder}
For the encoder, each atom type and edge type is first embedded in $\real^d$, then these feature are processed by $L$ layers of a graph neural network. We use the graph ConvNet technique introduced in \cite{bresson2017residual} to compute the hidden node and edge feature representations. Let $h_i \in \real^d$ and $e_{ij} \in \real^d$ denotes the feature vectors on vertex $i$ and edge $(i,j)$ of a graph. If the graph has $N$ vertices, then $h$ is an $N\times d$ matrice, and $e$ is an $N\times N\times d$ tensor.  The graph convolutional network will update $h$ and $e$ as follow
\begin{eqnarray}
(h^{\ell+1} , e^{\ell+1} ) = \textrm{GCN} (h^{\ell} , e^{\ell}) \label{eqgcn}
\end{eqnarray}
where
\begin{eqnarray}
h_i^{\ell+1} &= h_i^{\ell} + \textrm{ReLU} \Big( \textrm{BN} \Big( W_1^\ell h_i^{\ell} + \sum_{j\sim i} \eta_{ij}^{\ell} \odot W_2^\ell h_j^{\ell} \Big)\Big),\\
e_{ij}^{\ell+1} &= e^\ell_{ij} + \textrm{ReLU}\Big( \textrm{BN} \Big( V_1^\ell e_{ij}^{\ell} + V_2^\ell h^{\ell}_i + V_3^\ell h^{\ell}_j  \Big)\Big),
\end{eqnarray}
and
\begin{eqnarray}
\eta_{ij}^{\ell} = \frac{\sigma(e_{ij}^{\ell})}{\sum_{j'\sim i} \sigma(e_{ij'}^{\ell}) + \varepsilon },\end{eqnarray}
where $\eta_{ij}^{\ell}$ is a dense attention function, $\sigma$ and ReLU denote the standard sigmoid and ReLU non-linearities, and BN stands for batch normalization. We denote by $v \odot w$ the  component-wise multiplication between vectors $v$ and $w$. Each layer has a different set of parameters (i.e. the weights are not tied). Finally, a reduction step is applied in order to produces a vector $z$ of fixed size. The graph latent representation is given by the gated sum of edge features:
\begin{eqnarray}
z = \sum_{i,j=1}^N \sigma\left( Ae_{ij}^L + B h_i^L + Ch_j^L \right) \odot D e_{ij}^L,
\end{eqnarray}
where $z\in \real^k$ and $k$ is the latent vector dimension. The matrices $A$, $B$, $C$, $D$, $V$ and $W$ are the parameters to be learned.

\subsection{Atom generation}
The decoder first step is to generate a molecular formula. A molecular formula indicates the numbers of each type of atom in a molecule, with no information on bond structure. For example the molecular formula of carbon dioxide is $\textrm{CO}_3$, indicating that this molecule contains one carbon and three oxygens. A molecular formula can be seen as a simple ``bag-of-atom'' representation of a molecule. Let us assume for the sake of exposition that there are only 3 types of atoms in the set of possible atoms:
\begin{equation} 
\mathcal{A} = \{ \textrm{C, N, O} \}.
\end{equation} 
Then the molecular formula of carbon trioxide can be represented by the vector $v=[1,0,3]$ (1 carbon, 0 nitrogen, 3 oxygens). More generally, if we consider molecules with $m$ possible types of atom, the molecular formula can be represented by a vector with $m$ entries that contains the count of each type of atom. Since the molecular formula is represented by a vector of fixed size $m$, it can be easily produced by a fully connected neural network. Hence, the first step of the decoder is to feed the latent vector $z$ to a fully connected neural network, here a one-hidden-layer MLP, to produce a soft molecular formula:
\begin{eqnarray}
z_\textrm{soft-boa} = \textrm{MLP}_\textrm{boa}(z),
\end{eqnarray}
where $z_\textrm{soft-boa}$ is a $m\times r$ matrix, where $m$ is the number of atom types and $r$ is the largest possible molecule size in the training set. The molecular formula $z_\textrm{boa}\in \real^m$ is finally produced by taking the index of the maximum score along the second dimension of $z_\textrm{soft-boa}$. Once the molecular formula has been generated, the decoder will decide on how to connect each atom by generating the bonds between the atoms.

\subsection{Bond generation}
The decoder second step will take the bag-of-atoms vector $z_\textrm{boa}$ and the graph latent vector $z$ to assemble the atoms in a single pass. To do this, we start by creating a fully connected graph by connecting each atom in the molecular formula with one another. Each vertex of the fully connected graph receives a feature in $\real^d$ corresponding to the atom type via some embedding matrix, and each of the $N^2$ edges receives the same embedded feature vector $Uz$, where $z$ is the molecule latent vector, and $U$ is some learnable weight matrix. This fully connected graph is then processed by $L$ layer of the graph convolutional network described by Eq. \ref{eqgcn}, with new parameters for molecular decoding. The resulting feature vector $e_{ij}^{L}$ of the last convolutional layer can then be used to predict the type of bonds connecting atom $i$ to atom $j$ among possible types in 
\begin{equation} 
\mathcal B  = \{\textrm{None, Single, Double, Triple} \},
\end{equation}
where a bond type ``None'' corresponds to a no-bond between atoms.  A simple way of predicting the edge type is to use a MLP that classifies each of the vector $e_{ij}^{L}$ independently:
\begin{eqnarray}
s_{ij} = \textrm{MLP}_\textrm{edge}(e_{ij}^{L}),
\end{eqnarray}
where $s_{ij}\in \real^n$ is an edge score and $n$ is the number of bonds in $\mathcal B$. The edge type is eventually selected by taking the index of the maximum edge score. We will introduce a more sophisticated beam search strategy for edge type selection that leads to better results.

\subsection{Breaking the symmetry}
Consider the fully connected graph depicted in the top-right of Figure \ref{description}. At initialization, each of the 5 edges of the bond decoder has the exact same feature $Wz$, and each of the 3 carbon atoms has the same feature vector (the embedding vector of the carbon type). When this graph will be processed by the GCN, the features on the carbon atoms will not be able to differentiate from one another (as well as the features on the 3 edges connecting each carbon to the oxygen). In order to remedy to this symmetry problem, we introduce some positional features which allow to embed atoms of the same type into different vectors, and thus differentiate atoms of the same type.

\noindent
{\bf Positional features. }   Consider the chemical compound dichlorine hexoxide  which has molecular formula $\textrm{Cl}_2\textrm{O}_6$ (2 chlorines and 6 oxygens). Let assume that we have a natural way to order the atoms in the molecule so that the  8 atoms composing dichlorine hexoxide can be written as
  \begin{equation} \label{smiles}
  (\textrm{Cl},1) \;\; \;\;  (\textrm{Cl},2) \;\;\;\;   (\textrm{O},1) \;\;\;\;  (\textrm{O},2) \;\;\;\;   (\textrm{O},3) \;\;\;\;   (\textrm{O},4) \;\;\;\;   (\textrm{O},5) \;\; \;\;  (\textrm{O},6)
  \end{equation}
  where $(\textrm{O},3)$, for example, means ``$3^{\textrm{rd}}$ oxygen in the molecule".  We refer to the number 3 in this example as the ``positional feature". It allows us to distinguish this specific oxygen atom from the 5 other oxygen atoms. In order to obtain positional features, we need a consistent way to order the atoms appearing in a molecule (to be more precise, we need a consistent way to rank atoms compared to other atoms of the same type). In this work we simply order the atoms according to the position in which they appear in the canonical SMILES\footnote{SMILES stands for Simplified Molecular Input Line Entry System. } representation of the molecule (a SMILES is a single line text representation of a molecule).  To take an example, the notation $(\textrm{O},3)$ in (\ref{smiles}), means that this oxygen atom is the third one appearing in the SMILES representation of the dichlorine hexoxide compound. It is important to note that these position features contain some weak structural information about the molecule (this is a consequence of the algorithm that is used to compute a canonical SMILES). For example two carbon atoms $(\textrm{C},4)$ and $(\textrm{C},5)$ are very likely connected to each other. A carbon atom $(\textrm{C},1)$ is very likely at the beginning of a carbon chain. 
  
  Using  both atom type and positional feature allow us to build better embeddings for the atoms. 
  Let $M$ be an upper bond on the number of atoms contained in the molecule of interest, and let $m$ be the number of different types of atom. 
  To obtain an embedding of  $(\textrm{O},3)$, the ``$3^{\textrm{rd}}$ oxygen in the molecule",  we first concatenate a one-hot-vector in $\real^m$ representing O and a one-hot-vector in $\real^M$ representing the position feature 3.   The resulting vector, which is in $\real^{M+m}$,  is then  multiplied by a $d$-by-($M+m$) matrix in order to obtain an embedding in $\real^d$.

\subsection{Variational Auto-Encoder}
Finally, we use a VAE formulation in \cite{kingma2013auto} to improve the molecule generation task by ``filling'' the latent spac. The VAE requires to learn a parametrization of the molecular latent vector $z$ as 
\begin{eqnarray}
z = \mu + \sigma \odot \varepsilon, \quad \varepsilon \in \mathcal{N}(0,I),
\end{eqnarray}
where $\mu, \sigma$ are learned by the encoder with a reduction layer:
\begin{eqnarray}
f &= \sum_{i,j=1}^N \sigma\left( A_f e_{ij}^L + B_f h_i^L + C_f h_j^L \right) \odot D_f e_{ij}^L \quad \textrm{ for } \quad f=\mu,\sigma.
\end{eqnarray}
The total loss is composed of three terms, the cross-entropy loss for edge probability, the cross-entropy loss for bag-of-atoms probability, and the Kullback–Leibler divergence for the VAE Gaussian distribution:
\begin{eqnarray}
L = \lambda_{e}\sum_{ij} \hat{p}_{ij} \log p_{ij} + \lambda_a\sum_i \hat{q}_i \log q_i - \lambda_\textrm{VAE}\sum_k  (1+\log\sigma_k^2-\mu_k^2-\sigma_k^2).
\end{eqnarray}
Finally, no matching between input and output molecules is necessary because the same atom ordering is used (with the SMILES representation).

\subsection{Beam search}
The proposed one-shot decoder may not produce a chemically valid molecule because of a potential violation of atom valency. We use a greedy beam search technique to produce a valid molecule. The beam search is defined as follows. We start with a random edge. We select the next edge that (1) has the largest probability (or by Bernouilli sampling), (2) is connected to the selected edges, and (3) does not violate valency. When the edge selection ends then one molecule is generated. We repeat this process for a number $N_b$ of different random initializations, generating $N_b$ candidate molecules. Finally, we select the molecule that maximizes the product of edge probabilities or the chemical property to be optimized s.a. druglikeness, constrained solubility, etc.

\section{Experiments}
Our experimental setup follows the work of \cite{jin2018junction}.

{\bf Molecule dataset. } We use the ZINC molecule dataset \cite{irwin2012zinc}, which has 250k drug-like molecules, with up to 38 heavy atoms (excluded Hydrogen). The dataset is originally coded with SMILES. We use the open-source cheminformatics package Rdkit\footnote{http://www.rdkit.org/} to obtain  canonical SMILES representation.

{\bf Training. }  We trained the auto-encoder with mini-batches of 50 molecules of same size. The learning rate is decreased by 1.25 after each epoch if the training loss does not decrease by 1\%. The optimization process stops when the learning rate is less than $10^{-6}$. The average training time is 28 hours on one Nvidia 1080Ti GPU.

\begin{figure*}[h!]
\centering
\includegraphics[width=14cm]{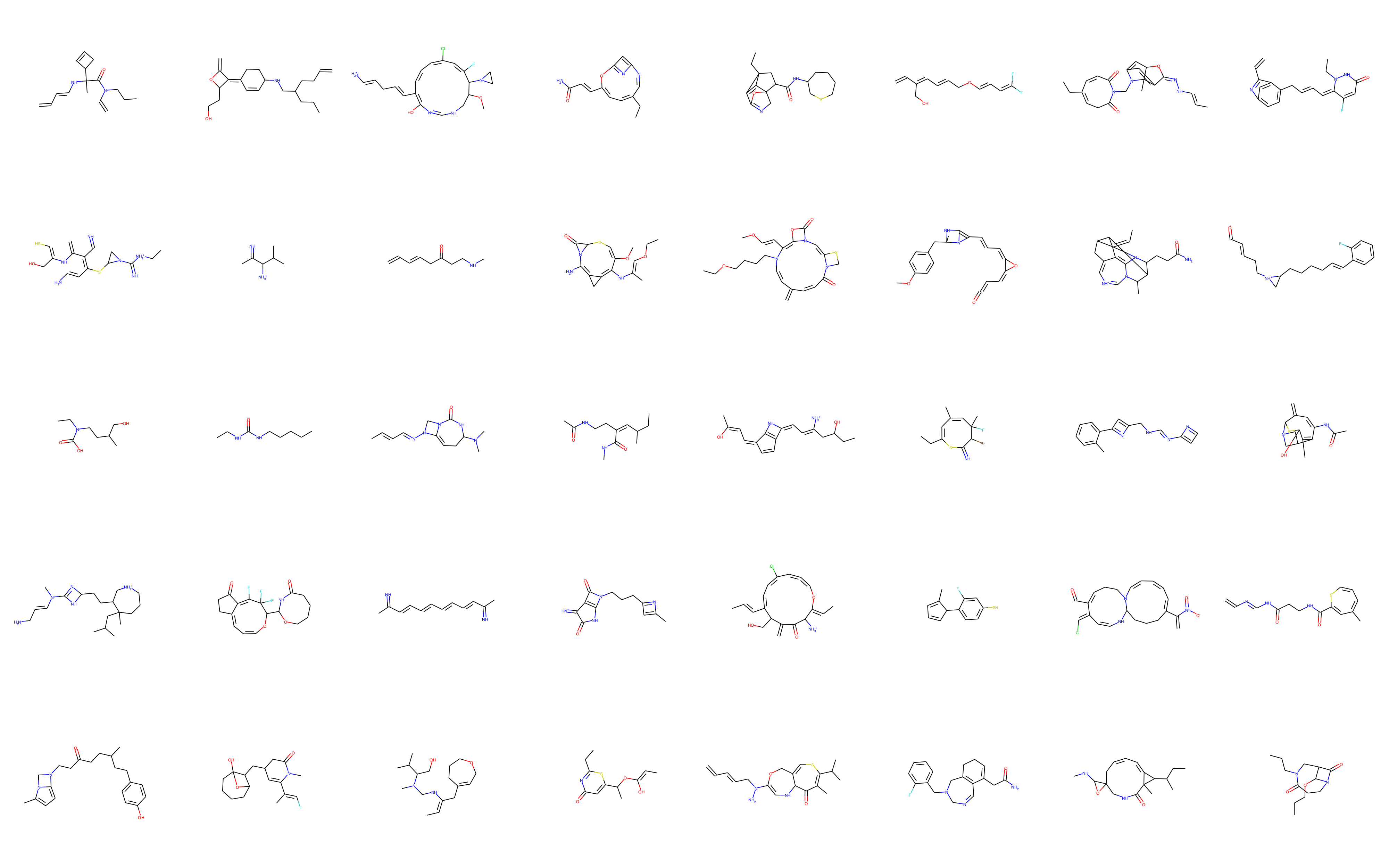}
\caption{Generated molecules sampled from $\mathcal{N}(0,I)$. 
}
\label{fig_sampling}
\end{figure*}

{\bf Molecule reconstruction, novelty and uniqueness. }  The first task is to reconstruct and sample molecules from the latent space. Table \ref{tab_recons} reports the reconstruction and valididty results. We improve the previous state-of-the-art reconstruction accuracy from 76.7\% to 90.5\% of \cite{jin2018junction}, with 100\% validity (no violation of atom valency) even for the molecules that were not correctly reconstructed. This is an improvement of almost 14\% upon previous works. We do not report the reconstruction value for the GAN approach of \cite{you2018graph}, as a GAN is not designed to auto-encode. To evaluate the novelty and uniqueness of the system, we sample 5000 molecules from the latent space from the prior distribution $\mathcal{N}(0,I)$ as in \cite{jin2018junction}. Table \ref{tab_sampling} presents the result. Our system does not simply memorize the training set, it is also able to generate 100\% of new valid molecules. Besides, all novel molecules are distinct to each other as the novelty measure is 100\% (percentage of molecules that are unique in the generated dataset). Figure \ref{fig_sampling} presents a few generated molecules.

\begin{table}[h!]
\centering
\begin{tabular}{ccc}
\hline
  \bf{Method} &   \bf{Reconstruction}  &   \bf{Validity} \\
  \hline
 CVAE, \cite{gomez2018automatic} & 44.6\%  & 0.7\% \\
 GVAE, \cite{kusner2017grammar} & 53.7\%  & 7.2\% \\
 SD-VAE, \cite{dai2018syntax} & 76.2\%  & 43.5\% \\
 GraphVAE, \cite{simonovsky2018graphvae} & -   & 13.5\% \\
 JT-VAE, \cite{jin2018junction} & 76.7\%  &  \bf{100.0\%} \\
 GCPN, \cite{you2018graph} & -  & - \\
  \hline
 OURS & \bf{90.5\%}  & \bf{100.0\%}\\ 	
   \hline
\end{tabular}
\vspace{0.15cm}
\caption{Encoding and decoding of the 250k ZINC molecules.}
\label{tab_recons}
\end{table}

\begin{table}[h!]
\centering
\begin{tabular}{ccc}
\hline
  \bf{Method} &   \bf{Novelty}  &   \bf{Uniqueness} \\
  \hline
  JT-VAE,  \cite{simonovsky2018graphvae} &  \bf{100.0\%}  &  \bf{100.0\%} \\
 GCPN, \cite{you2018graph} & -  & - \\
  \hline
 OURS & \bf{100.0\%}  & \bf{100.0\%}\\ 	
   \hline
\end{tabular}
\vspace{0.15cm}
\caption{Sample 5000 molecules from prior distribution.}
\label{tab_sampling}
\end{table}

\newpage

{\bf Property Optimization. }  The second task is to produce new molecules with optimized desired chemical property. We follow the experiment of \cite{jin2018junction,kusner2017grammar} and select the target property to be the octanol-water partition coefficients (logP) penalized by the synthetic accessibility (SA) score and number of long cycles. To perform the molecular optimization, we train our VAE to simultaneously auto-encode the training molecule and the target chemical property. For this, we add a MLP layer after the graph convolutional encoder to predict the chemical property and a L2 regression loss that penalizes bad property prediction.

\begin{figure*}[h!]
\centering
\includegraphics[width=14cm]{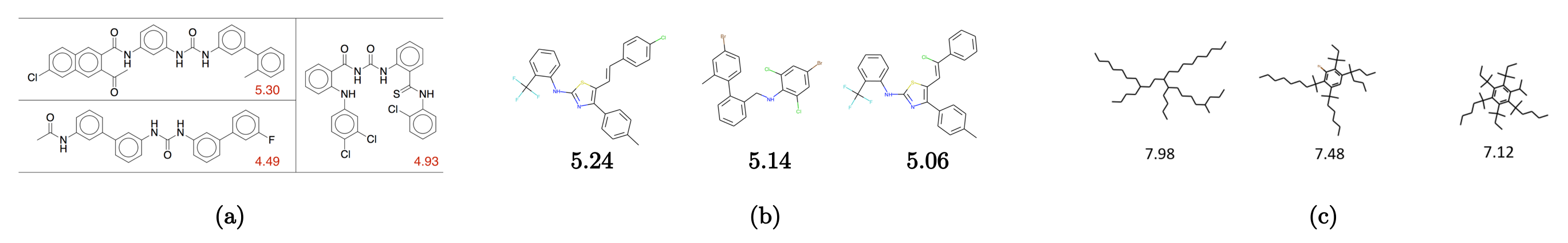}
\caption{Generated top-3 molecules. (a) \cite{jin2018junction}, (b) Ours and (c) \cite{you2018graph}.
}
\label{fig_top3}
\end{figure*}

We perform gradient ascent w.r.t. the chemical property to optimize in the latent space, then decode the molecule and compute the target property. We optimize the molecules that have the top 100 property values in the training set. Table \ref{tab_top3} reports the top-3 molecules from our model and the literature. Our VAE model does slightly better in average, 5.14 vs. 4.90, than the previous state-of-the-art VAE model of \cite{jin2018junction}. However, the RL model of \cite{you2018graph} does significantly better than all VAE techniques with a mean value of 7.88. This is expected as RL approaches can extrapolate new data that can be outside the statistics of the training set, whereas VAE approaches can only interpolate data inside the training statistics. Figure \ref{fig_top3} presents the top-3 optimized molecules.

\begin{table}[h!]
\centering
\begin{tabular}{ccccc}
\hline
  \bf{Method} &   \bf{1st }  &   \bf{2nd } &   \bf{3rd } &   \bf{Mean }    \\
  \hline
  \vspace{-0.4cm}\\
  ZINC & 4.52 & 4.30 & 4.23 & 4.35   \\   
  \vspace{-0.4cm}\\
    \hline
 CVAE, \cite{gomez2018automatic} & 1.98 & 1.42 & 1.19 & 1.53   \\   
 GVAE, \cite{kusner2017grammar} & 2.94 & 2.89 & 2.80 & 2.87 \\
 SD-VAE, \cite{dai2018syntax} & 4.04 & 3.50 & 2.96 & 3.50 \\
 JT-VAE, \cite{jin2018junction} &  5.30 & 4.93 & 4.49 & 4.90 \\
  \hline
    \vspace{-0.4cm}\\
  OURS (VAE+SL)  & 5.24 &  5.10 & 5.06 & 5.14  \\
    \vspace{-0.4cm}\\
    \hline
  \vspace{-0.4cm}\\
GCPN (GAN+RL), \cite{you2018graph} & \bf{7.98} & \bf{7.85} & \bf{7.80} & \bf{7.88} \\
  \vspace{-0.4cm}\\
  \hline
  \end{tabular}
\vspace{0.15cm}
\caption{Generative performance of the top three molecules for Penalized logP (logP-SA-Cycle) trained on VAE+SL and GAN+RL.}
\label{tab_top3}
\end{table}

{\bf Constrained Property Optimization. }  The third task is to generate novel molecules with optimized  chemical property while also constraining molecular similarity between the original molecule and the generated molecule. This is important when we want to optimize the property of a promising molecule in drug discovery and materials science. We follow again the experiment setup of \cite{jin2018junction,kusner2017grammar}. The goal is to maximize the constrained solubility of the 800 test molecules with the lowest property value. We report in Table \ref{fig_cons_opt} the property improvements w.r.t. the molecule similarity $\delta$ between the original molecule and the generated molecule. Our model outperform the previous state-of-the-art VAE model of \cite{jin2018junction} and the RL model of \cite{you2018graph} for the property improvement. The RL model outperforms all VAE models for the success rate (a model is successful when it is able to generate a new molecule with the given molecular distance $\delta$). Figure \ref{fig_cons_opt} presents constrained optimized molecules w.r.t. the molecular distance.

\begin{table}[h!]
\centering
\resizebox{\textwidth}{!}{%
\begin{tabular}{cccccccccccc}
\cline{2-4} \cline{6-8} \cline{10-12} 
& \multicolumn{3}{c}{JT-VAE,  \cite{jin2018junction}  (VAE+SL)} & & \multicolumn{3}{c}{GCPN,  \cite{you2018graph}  (GAN+RL)} && \multicolumn{3}{c}{OURS (VAE+SL)} \\
\cline{2-4} \cline{6-8} \cline{10-12} 
$\delta$ & Improvement & Similarity & Success & & Improvement & Similarity & Success 
& & Improvement & Similarity & Success\\
\cline{2-4} \cline{6-8} \cline{10-12} 
0.0 & 1.91 $\pm$ 2.04 & 0.28 $\pm$ 0.15 & 97.5\% & & 4.20 $\pm$ 1.28 & \bf{0.32 $\pm$ 0.12} & \bf{100.0\%} &&   \bf{5.24 $\pm$ 1.55} & 0.18 $\pm$ 0.12 & \bf{100.0\%}\\
0.2 & 1.68 $\pm$ 1.85 & 0.33 $\pm$ 0.13 & 97.1\% & &4.12 $\pm$ 1.19 & \bf{0.34 $\pm$ 0.11} &\bf{100.0\%} &&  \bf{4.29 $\pm$ 1.57} & 0.31 $\pm$ 0.12 & 98.6\%\\
0.4 & 0.84 $\pm$ 1.45 & \bf{0.51 $\pm$ 0.10} & 83.6\% & & 2.49 $\pm$ 1.30 & 0.47 $\pm$ 0.08 & \bf{100.0\%}  &&\bf{3.05 $\pm$ 1.46} & \bf{0.51 $\pm$ 0.10} & 84.0\%\\
0.6 & 0.21 $\pm$ 0.71 & \bf{0.69 $\pm$ 0.06} & 46.4\% & & 0.79 $\pm$ 0.63 & 0.68 $\pm$ 0.08 & \bf{100.0\%} && \bf{2.46 $\pm$ 1.27} & 0.67 $\pm$ 0.05 & 40.1\%\\
\cline{2-4} \cline{6-8} \cline{10-12} 
\end{tabular}
}
\vspace{0.15cm}
\caption{Molecular optimization with constrained distances.}
\label{fig_cons_opt}
\end{table}

\begin{figure*}[h!]
\centering
\includegraphics[width=14cm]{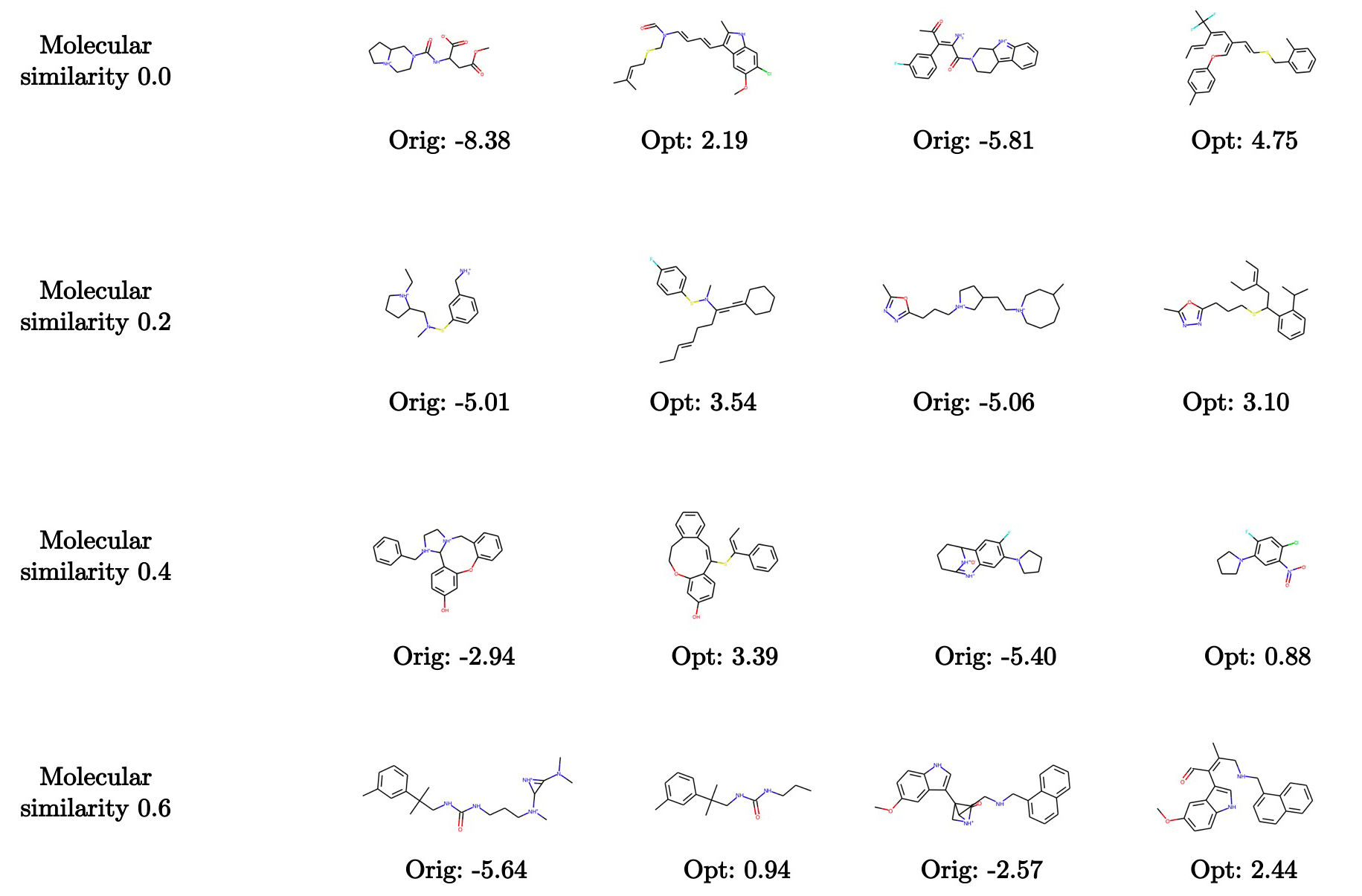}
\caption{Molecule modification that yields to chemical improvement constrained by molecular similarity. ``Orig'' means original molecule and ``Opt'' means optimized molecule. 
}
\label{fig_cons_opt}
\end{figure*}

\section{Conclusion}
We introduce a simple and efficient VAE model for the molecule generation task. Our decoder generates the molecular formula and the bond structure in one-shot, which can be a faster alternative to autoregressive models. To the best of our knowledge, this is also the first time that beam search is used for improving the molecule generation task. This is also attractive as beam search can be highly parallelized, as for natural language processing systems. Overall, the proposed technique is simpler to implement w.r.t. previous autoregressive VAE techniques such as \cite{jin2018junction}, which make use of chemical handcrafted features such as ring and tree structure or a molecular grammar in \cite{kusner2017grammar}. We report the highest reconstruction rate for the ZINC dataset. We do not beat the RL technique of \cite{you2018graph} for optimizing absolute molecular property, but we report the best property improvement results when the optimization is constrained by the distances between the original molecule and the new optimized one. This demonstrates that the proposed VAE is able to learn a good latent space representation of molecules. Future work will explore a RL formulation of the proposed non-autoregressive technique.

\section{Acknowledgement}
Xavier Bresson is supported in part by NRF Fellowship NRFF2017-10.

\bibliographystyle{plainnat}
\bibliography{mol_paper_bib}

\end{document}